\documentclass{article}

\usepackage{microtype}
\usepackage{graphicx}
\usepackage{subfigure}
\usepackage{booktabs} % for professional tables

\usepackage{hyperref}

\usepackage[accepted]{icml2023}

\usepackage{amsmath}
\usepackage{amssymb}
\usepackage{mathtools}
\usepackage{amsthm}

\usepackage[capitalize,noabbrev]{cleveref}

\theoremstyle{plain}
\newtheorem{theorem}{Theorem}[section]

\theoremstyle{definition}

\theoremstyle{remark}

\usepackage[textsize=tiny]{todonotes}

\begin{document}

\twocolumn[
\icmltitle{A Fast Learning-Based Surrogate of Electrical Machines using a Reduced Basis}



\icmlsetsymbol{equal}{*}

\begin{icmlauthorlist}
\icmlauthor{Alejandro Ribés}{comp,sin}
\icmlauthor{Nawfal Benchekroun}{comp}
\icmlauthor{Théo Delagnes}{comp}

\end{icmlauthorlist}

\icmlaffiliation{comp}{EDF Lab Paris-Saclay, Palaiseau, France}
\icmlaffiliation{sin}{Industrial AI Laboratory SINCLAIR, France}

\icmlcorrespondingauthor{Alejandro Ribés}{alejandro.ribes@edf.fr}

\icmlkeywords{Machine Learning, ICML}

\vskip 0.3in
]



\printAffiliationsAndNotice{}  % leave blank if no need to mention equal contribution

\begin{abstract}
  A surrogate model approximates the outputs of a solver of Partial Differential Equations (PDEs) with a low computational cost. In this article, we propose a method to build learning-based surrogates in the context of parameterized PDEs, which are PDEs that depend on a set of parameters but are also temporal and spatial processes.
  Our contribution is a method hybridizing the Proper Orthogonal Decomposition and several Support Vector Regression machines.
  This method is conceived to work in real-time, thus aimed for being used in the context of digital twins, where a user can perform an interactive analysis of results based on the proposed surrogate. We present promising results on two use cases concerning electrical machines. These use cases are not toy examples but are produced an industrial computational code, they use meshes representing non-trivial geometries and contain non-linearities.
\end{abstract}

\section{Introduction}
\label{sec:introduction}

In the context of numerical simulation, a surrogate model approximates the outputs of a solver with a low computational cost. Solvers of differential equations, for instance, those based on finite element methods, often require long runs. Thus, they are not well-suited for real-time applications. In the last five years, the use of machine learning for constructing surrogate models has gained a lot of attention from industry and academics. These surrogates learn from simulation results and/or experimental data. Numerous methods for building surrogates have been proposed, in section \ref{sec:relatedWork} we discuss some of them, which we find representative of the variety of learning-based surrogates currently being developed. However, after analyzing the technical literature, we found two main limitations of the current state-of-the-art. First, the majority of use cases correspond to toy examples. For instance, the 1D Burgers equation is a popular choice. In general, the test use cases found in the literature are far in geometric complexity and size from what computation codes deal with in the current industrial practice. Second, the question of the eventual scaling of the method or its real-time capabilities is not often studied.

In this article, we propose a method to build surrogates in the context of parameterized Partial Differential Equations (PDEs), which depend on a set of parameters, that may represent boundary or initial conditions, materials laws, or other physical properties. In the general case, constructing surrogates for parameterized PDEs is a complex task.
Our objective is to develop a learning-based surrogate that can be used to power a Digital Twin. This surrogate should be executed on a computer sufficiently fast to allow interactive exploration of the parameters of the underlying PDEs but also on space and time. This means that the surrogate should reconstruct a scalar or vectorial field at any time step and for any combination of the parameters. This is a state-of-the-art problem, especially when this task is performed in real time.

We have designed a method hybridizing the Proper Orthogonal Decomposition (POD, see chapter 11 of \cite{brunton_kutz_2019}) and several Support Vector Regression machines (SVR, see \cite{drucker1996svr}), where the Reduced Basis obtained by the POD is used to facilitate the training of our Machine-Learning system.
We have designed a two-step simple method which is very fast at inference due to its simplicity.
The method deals with geometric domains represented by non-regular meshes. Concerning the temporal discretization, it takes a \emph{direct time} approach, which means it produces the state corresponding to the time step provided as input. This approach, which avoids iterating over time, is faster than autoregressive methods and thus well-adapted to real-time applications. The output of the proposed surrogate is an array containing a whole vectorial or scalar field, corresponding to the specified input set of parameters.

This article is structured as follows. Section \ref{sec:relatedWork} discusses relevant related work. Section \ref{sec:method} describes our proposed method for constructing learning-based surrogates. Section \ref{sec:use-cases} introduces two use cases, concerning electric machines. These use cases are not toy examples but are produced by an industrial computational code. They use meshes representing non-trivial geometries
and contain non-linearities. Results are shown in Section \ref{sec:results}. Section \ref{sec:realTime} discusses the real-time capabilities of the proposed method. Section \ref{sec:conclusion} concludes the article. Finally, a bibliography section is given, followed by Appendix \ref{sec:appendix1} which contains a mathematical proof for the inference error upper bound of the proposed method.

\section{Related Work}
\label{sec:relatedWork}

Reducing the computational complexity of numerical simulations is not a new subject. The Reduced Order Model (ROM) community has traditionally tackled this problem. A ROM can be considered a type of surrogate model.  Constructing surrogates for parameterized PDEs (the problem of this article) is, in the general case, a very complex task \cite{quarteroni2016reducedBasis}. One of the most popular techniques for building ROMs is the Proper Orthogonal Decomposition (POD, see chapter 11 of \cite{brunton_kutz_2019}), which we use in our work. On the other hand, building surrogates of simulations by using machine learning techniques is a recent research domain. From the seminal work of Raissi et al. \cite{raissi2019physics}, which introduced the so-called physics-informed neural networks
(PINNs), numerous techniques have been developed. 
Most of the learning-based surrogates found in the literature are trained in a supervised manner from simulation data. PINNs could stand as an exception, as they can be trained unsupervised.
This is because PINNs are trained by minimizing the residual error of the PDE at random collocation points \cite{raissi2019physics,sirignano2018dgm,wandel2020learning}.
 But  PINNs also benefit from training with simulation data \cite{krishnapriyan2021characterizing,lucor2022simple}. Our method is also a supervised learning method that learns from the results of an ensemble of pre-executed numerical simulations.

Building learning-based surrogates still presents several important challenges, which are associated with different aspects of the numerical simulations. In the following, we discuss some of these aspects. 

\textbf{Spatial discretization}. When the meshes supporting the numerical simulations are regular, they can be seen as images and convolutional networks can be employed successfully \cite{zhu2019PI-NNSurrogate, ronneberger2015u, kasim2021building}. For instance, U-Net architectures are employed in
\cite{thuerey2020deep,wang2020towards}, or auto-encoders in \cite{kim2019deepFluids}.
For non-regular meshes Graph Neural Networks (GNNs) \cite{bronstein2017geometric,battaglia2018relational,brandstetter2021message} have been used. As an example, Deep-Mind introduced a GNN-based framework for learning mesh-based simulations \cite{pfaff2020learning,sanchez2020learning}.
Our method also supports the use of non-regular meshes.

\textbf{Time discretization}. Not only does the space discretization influence the design of the architectures, but the handling of the time dimension also distinguishes \emph{autoregressive} from \emph{direct} models. Autoregressive models mimic the iterative process of traditional solvers, where the current state is used as input to predict the next one. One of the challenges for autoregressive models is the error accumulation along a trajectory, leading to various mitigation strategies \cite{brandstetter2021message,pfaff2020learning,takamoto2022pdebench}. Some researchers have also integrated the time dimension using recurrent architectures \cite{tang2020deep} or attention mechanisms \cite{li2023transformer}. Direct models produce the state corresponding to the time step provided as input. 
PINNs are an example of direct models, our proposed method also uses this approach.

\textbf{Respecting physical laws}. From \cite{karniadakis2021pinns}, three main ways allow a learning system to represent correct physical laws: observational, inductive, and learning biases. PINNs \cite{raissi2019physics} focus on the learning biases by introducing physical constraints on the loss function. Techniques such as GNNs \cite{pfaff2020learning,sanchez2020learning} focus on
designing specialized neural network architectures that implicitly embed prior knowledge about the problem (in this case information about the spatial domain) thus using inductive biases. Observational bias is the fact that the data used for training the system contains relevant and correct physical information. In this article, we use observational bias by preparing designs of experiences that execute an ensemble of meaningful numerical simulations \cite{santner2003design}.

\textbf{Operators}. PDEs are theoretically treated in the context of operators, which are maps between infinite-dimensional function
spaces. However, neural networks learn mappings between finite-dimensional Euclidean spaces or finite sets \cite{kovachki2023neuralNvidia}. Efforts exist to create the so-called Neural Operators, which are neural networks aiming to map between functional spaces. Examples of this approach are techniques such as DeepONets \cite{lu2021deepONet}, Graph Kernel Networks for PDEs \cite{li2020neuralOperator}, or Fourier neural operators \cite{li2020fourier}. Neural operators aim at building surrogates of parametrized PDEs. Our method cannot be strictly considered a Neural Operator but it certainly maps from a parameter's space to a discretized functional space.

In summary, the method proposed in this article aims at building surrogates of parametrized PDEs. It is a direct time method that supports irregular meshes. The method learns from the results of an ensemble of pre-executed numerical simulations, thus using observational bias. A difference with the current literature is that its focus is not only on the quality of the reconstruction but also on the speed of the inference.

\section{Proposed Method}
\label{sec:method}

This section presents the technical details of the conceived algorithm that combines model reduction via the POD and \textit{support vector regression} (SVR).  It consists of two main steps: first, a reduced basis is found using the POD, and second several SVRs are trained. We remark that before these two steps, an ensemble of $N$ simulations should be run in order to create the dataset necessary for the learning. This step is not specific to our method, any learning-based surrogate needs a dataset to be trained on.

\subsection{Finding a Reduced Basis}
\label{sec:findRB}

We obtain a Reduced Basis using the Proper Orthogonal Decomposition (POD), see chapter 11 of \cite{brunton_kutz_2019} or chapter 6 of \cite{quarteroni2016reducedBasis}.
The fundamental step of the POD is the application of a Singular Value Decomposition \cite{strang1993introduction} to a so-called snapshot matrix. Thus we define here how we construct this matrix for parameterized problems. 

\textbf{Snapshot matrix for a parametric problem:} When dealing with parametric and spatio-temporal problems, we pre-compute an ensemble of $N$ spatio-temporal simulations. Each simulation contains $t$ times steps and has a different vector of parameters $\lambda$. Afterward, we can build the snapshot matrix as follows: 

\begin{equation}
\label{eq:snapshot}
\resizebox{1.0\hsize}{!} 
{$ X = \left[
  \begin{array}{ccccccc}
    \vrule  &   & \vrule  &        & \vrule & & \vrule\\
    X_{\lambda_1, t_1}    & \ldots &X_{\lambda_1, t_T}  & \ldots & X_{\lambda_{N}, t_1} & \ldots & X_{\lambda_{N}, t_T}   \\
    \vrule & & \vrule  &    & \vrule  &  &\vrule
  \end{array}
\right] $} 
\end{equation}

where each column $X_{\lambda_i,t_j}$ of $X$ contains a vector $x_{i,j}$ indexed by $(\lambda_i, t_j)$.
Thus, discretized and linearized fields, representing a physical quantity provided by the solver, are contained in each vector $x_{i,j}$.  We can build a \textit{matrix of snapshots} $X$ of size $(n, m)$ by concatenating these vectors. Note that $n$ is equal to the number of mesh cells (or nodes) used by the simulations multiplied by the number of components of the field (one for a scalar field, 2 or 3 for vectorial fields); a column of the matrix contains a spatially discretized field with its components stacked inside.
The matrix has $m=NT$ columns, where $N$ is the number of pre-computed simulations and $T$ is the number of time steps.

\textbf{SVD:} A description of the singular value decomposition (SVD) can be found in any introductory linear algebra book, such as \cite{strang1993introduction}. Let $X \in \mathbb{R}^{n \times m}, U \in \mathbb{R}^{n \times n}, \Sigma \in  \mathbb{R}^{n \times m}, V \in \mathbb{R}^{m \times m} $ the SVD of X is the decomposition $ X = U\cdot\Sigma\cdot V^T$, where $U$ and $V$ are unitary matrices and $\Sigma$ is a diagonal matrix containing the singular values of $X$, which are ordered by decreasing value.

Applying a Singular Value Decomposition to the snapshot matrix allows for finding an orthogonal basis. However, this basis is of the same size as the original non-transformed problem. The key to finding a \textbf{Reduced Basis} (RB) comes from the fact that not all the principal components need to be kept. Keeping only the first $r$ principal components, produced by using only the first $r$ eigenvectors, gives the truncated transformation. The value $r$ is typically found by looking at the accumulated energy, which is defined by:

\begin{equation}
E = \frac{\sigma_1+ \sigma_2+ ... + \sigma_r}{\sigma_1+ \sigma_2+ ... + \sigma_m}
\label{eq:sigs}
\end{equation}

where the $\sigma_i$ ($i=1...m$) values are the diagonal elements of the matrix $\Sigma$. Once the value $r$ is chosen we obtain the following approximation of the matrix X: 

\begin{equation}
X^r = U^r\cdot\Sigma^r\cdot V^{Tr}.
\label{eq:reducedSVD}
\end{equation}

\subsection{Training the SVRs}

\begin{figure}[hb]
  \centering
  \begin{tikzpicture}[scale=0.7]
    \draw[black, top color=green!20, bottom color=green!60, rounded corners](4,1.3) rectangle (7.6,-0.5);
    \node(analysis) at (5.8,0.3) {$SVR$};
    \draw[black, ->] (7.6,0.3) -- (8.6,0.3);
    \node[text width=2cm, style={align=center}] at (9,0.3) {$\hat{c}_i$};
    \node (simu2) at (2.5,0.8) {$t$};
    \node (simu1) at (2.5,0.0) {$\lambda$};
    \draw[black, ->] (3,0.8) -- (3.9,0.8);
    \draw[black, ->] (3,0.0) -- (3.9,0.0);
  \end{tikzpicture}

  \caption{A SVR takes as input a time step $t$ and a vector of parameters $\lambda$. It outputs a prediction $\hat{c}_i$, corresponding to the i-th coefficient on the reduced space, $i=1...r$.}
\vspace{-0.3cm}
  \label{SVR-in-out}
\end{figure}
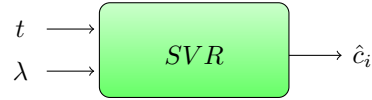

The SVR being a supervised learning algorithm, it is necessary to constitute $(input, output)$ pairs for its training. In figure \ref{SVR-in-out}, we depict what a single SVR takes in and out. The SVR accepts $(t, \lambda)$ as inputs, where $t$ is a time step and $\lambda$ is a vector of parameters. The SVR outputs a prediction $\hat{c}_i$, corresponding to the i-th coefficient on the reduced space, $i=1...r$.

\textbf{Preparing for the training phase}. 
Our idea is to project the snapshot matrix $X$ into the reduced space $C$. For this, we use $ U^r$ from equation \ref{eq:reducedSVD} but we call this projection matrix $U_{POD}^r$ to reinforce the fact that it is obtained by the POD.
Consequently, the operation $C = U_{POD}^r \cdot X$ projects the snapshot matrix on the reduced space. However, we need not only a matrix for training but also a matrix for validation. The matrix $X$ is therefore subdivided into $X_{train}$ and $X_{val}$, thus 

\begin{equation}
X = 
\left[
  \begin{array}{ccc}
    X_{train}  & \vrule  &   X_{val} 
  \end{array}
\right]
\end{equation}

We can then construct the corresponding matrices of coefficients   
$C = \left[ ~ C_{train} ~  \vrule    ~ C_{val} ~ \right]$
by matrix product:

\begin{equation}
C = 
\left[
  \begin{array}{ccc}
    U_{POD}^r \cdot X_{train}  & \vrule  &  U_{POD}^r \cdot X_{val} 
  \end{array}
\right]
\end{equation}

At this point, we have defined how to form the training and validation sets (matrices in this case) for the outputs of the SVR. In figure \ref{SVR-in-out}, we observe that the SVR accepts $(t, \lambda)$ as inputs. These inputs can be coded in the following matrix: 

\begin{equation}
x = 
\left[
  \begin{array}{ccccccccc}\label{snapshot}
    t_1  & t_2& \ldots & t_P  &   \ldots     & t_1 & t_2&  \ldots & t_P\\
    \vrule  & \vrule &  & \vrule  &        & \vrule & \vrule &  & \vrule\\
    \lambda_1    & \lambda_1& \ldots &\lambda_1  & \ldots & \lambda_{N} & \lambda_{N} &\ldots & \lambda_{N}   \\
    \vrule & \vrule&  & \vrule  &        & \vrule & \vrule &  &\vrule
  \end{array}
\right]
\end{equation} 
\vspace{0.2cm}

where each column $X_{\lambda_i,t_j}$ of $X$ (in equation \ref{eq:snapshot}) is associated with a vector $x_{i,j}=(\lambda_i, t_j)$.
Thus the matrix $x$ encodes the time and parameters in exactly the same way as $X$.
Similarly, $x$ is therefore subdivided into $x_{train}$ and $x_{val}$, thus $x = 
\left[
  \begin{array}{ccc}
    x_{train}  & \vrule  &   x_{val} 
  \end{array}
\right]$. Now it is possible to constitute the training and validation datasets which are respectively $(x_{train}, C_{train})$ and $(x_{val}, C_{val})$.

\textbf{Training of $r$ SVRs} 
The second stage of the training phase is to train $r$ \textit{SVRs}, one for each element of the vector $C_ {train}$. For example, the first \textit{SVR} is trained to predict the first element of $C_ {train}$ from $x_ {train}$. In other words, we can say that the i-th SVR is led to predict the value of the  \textit{parametro-temporal} coefficients of the i-th mode in the reduced space generated by $U_ {pod}^r$, from the parameter values contained in $x_ {train}$. We note that it is in general necessary to center and reduce the training and validation data sets before training the \textit{SVR}s.

\subsection{Tuning the hyperparameters of the SVRs }

The SVR method contains several hyperparameters that should be tuned. In this work, we have used the SVR implementation of Scikit-Learn \cite{scikit-learn} and performed tuning for three important parameters:
\begin{itemize}
    \item Epsilon in the epsilon-SVR model. It specifies the epsilon-tube within which no penalty is associated with the training loss function with points predicted within a distance epsilon from the actual value.
    \item The regularization parameter associated with a squared penalty term.
    \item We decided to fix the kernel used in the algorithm, which is a Gaussian. Thus, the standard deviation $\sigma$ of the Gaussian must be tuned.
\end{itemize}

We use the Optuna package \cite{akiba2019optuna} for tuning the above-presented parameters. We specifically use Optuna's implementation of the Tree-structured Parzen estimator (TPE) algorithm \cite{bergstra2011algorithms}, a Bayesian optimization method widely used in recent parameter tuning frameworks. 

We remark that SVRs can take several scalar inputs but they generate a unique scalar output. This implies that our method should train $r$ SVRs, where $r$ is the dimension of the reduced space. 
Thus a question arises concerning the $3r$ parameters to be optimized: is each SVR going to be treated independently, or can the $3r$ parameters be jointly tuned? We have found that, when jointly optimizing for the worst validation error, an upper bound on the error of the reconstructed fields exists.

We have proven that the reconstruction error of the physical field $\| \mathbf{X}_p-\hat{\mathbf{X}}_p \|_2$, computed at each cell $p\leq n$, has an upper bound that is linearly proportional to the highest validation error of all $r$ SVR machines (the whole prove is given in Appendix \ref{sec:appendix1}). Thus:

$$\| \mathbf{X}_p-\hat{\mathbf{X}}_p \|_2 \leq  K_p.\mathbf{e}$$

where $K_p$ is a positive real number, and $\mathbf{e}$ is the error. The constant $K_p$ comprises terms of the projection matrix $U_{POD}^r$ and standardization elements: $S_i$ is the standard deviation and $E_i$ is the mean of the coefficient vectors $C_i$. It is written as $K_p =  \sqrt{ A + 2 B}$ where:

$$A = \sum_{k=1}^{k=r}(S_k. U_{POD_{pk}}^r)^2$$

$$B = \sum_{1\leq h<l \leq r}\lvert S_h. U_{POD_{ph}}^r\rvert . \lvert S_l. U_{POD_{pl}}^r\rvert$$

\section{Description of the use cases}
\label{sec:use-cases}

We present two electrical machines use cases generated by Code$\_$Carmel (code-carmel.univ-lille.fr), which is an industrial computational code. Table \ref{table:case-comparisson} shows a comparison of these use cases concerning: the number of cells in the mesh, number of time steps, number of runs performed by the design of experiments, and size of the data when stored on disk (compressed).

\begin{table}[tb!]
\caption{Comparison of the two use cases concerning: the number of cells in the mesh, number of time steps, number of runs performed by the design of experiments, and size of the data stored on disk.}
\label{table:case-comparisson}
\vskip 0.15in
\begin{center}
\begin{small}
\begin{sc}
\begin{tabular}{lcccr}
\toprule
Use Case & Cells & Steps & Runs & Storage \\
\midrule
Transformer & 88,072 & 40 & 327 & 26G \\
Induction   & 30,047 & 10 & 750 & 5.3G \\
\bottomrule
\end{tabular}
\end{sc}
\end{small}
\end{center}
\vskip -0.1in
\end{table}

\subsection{An induction plate}
\label{subsec:use-case-Induction}

In the first use case, we are interested in a magneto-quasistatic phenomenon by the modeling of a magnetic electric current inductor. This could represent several applications, such as an induction hotplate in a kitchen or the induction charger of a mobile phone.
The inductor is a coil of 700 turns at which a current is injected
 with a sinusoidal frequency $f$. The intensity of the current in the coil is calculated from the electrical voltage
 $U$ imposed and a characteristic resistance of the inductor $R_{ind}$ via a standard circuit equation.
The plate, made of copper, has an electrical conductivity of
$\sigma = 5 \cdot 10^{7} S \cdot m^{-1}$.

\begin{figure}[h]
    \centering
    \includegraphics[width=.8\columnwidth]{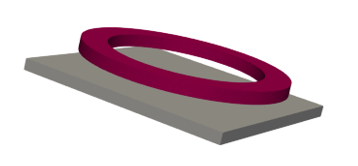}
    \caption{Mesh of the induction plate use case.}
    \label{fig:use-case-induction}
  \end{figure}

\textbf{Design of experiments.} Four parameters are considered in this use case: $f$, $U$, $R_{ind}$, and the time $t$. We run 750 simulations, each one corresponding to a parameter set. The values of $f$, $U$, $R_{ind}$ are sampled from a non-informative uniform distribution. The sampling of $t$ is given by the computational code.

\textbf{Why this use case?}
The choice of this use case is motivated by two factors.
First, the dynamics of the problem impose a time resolution of the finite elements model which can be computationally expensive. Then, we are interested in the accuracy of the proposed method for time-dependent problems, as it might provide large calculation speedups.
Second, the calculation of the source current through a circuit equation impacts the dynamical evolution of the problem. Therefore, we aim to evaluate the capacity of the proposed method to adapt not only to the electromagnetic dynamics but also to changes in the circuit equation parameters ($f$, $U$, $R_{ind}$).

\subsection{A three-phase transformer}
\label{subsec:use-case-Transfo3D}

A power network is a complex system, with a large variety of voltage levels at generation, transmission, and distribution points. In this context, transformers are vital components, as they allow transitioning from one voltage level to the other, ensuring the transmission of electric energy through the network. They are static devices, consisting of one or multiple windings and a magnetic core, used extensively by electric companies in all types of production sites. For these reasons, power transformers are the subject of numerous numerical simulations for diagnosis and design review purposes, and surrogate models have been proposed in recent years to reduce computational costs \cite{henneron2014model}. 

\begin{figure}[h]
    \centering
    \includegraphics[width=.7\columnwidth]{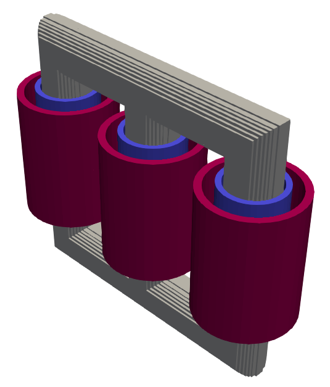}
    \caption{Mesh of the three-phase transformer use case.}
    \label{fig:use-case-transformer}
  \end{figure}

\begin{figure}[h]
    \centering
    \includegraphics[width=.9\columnwidth]{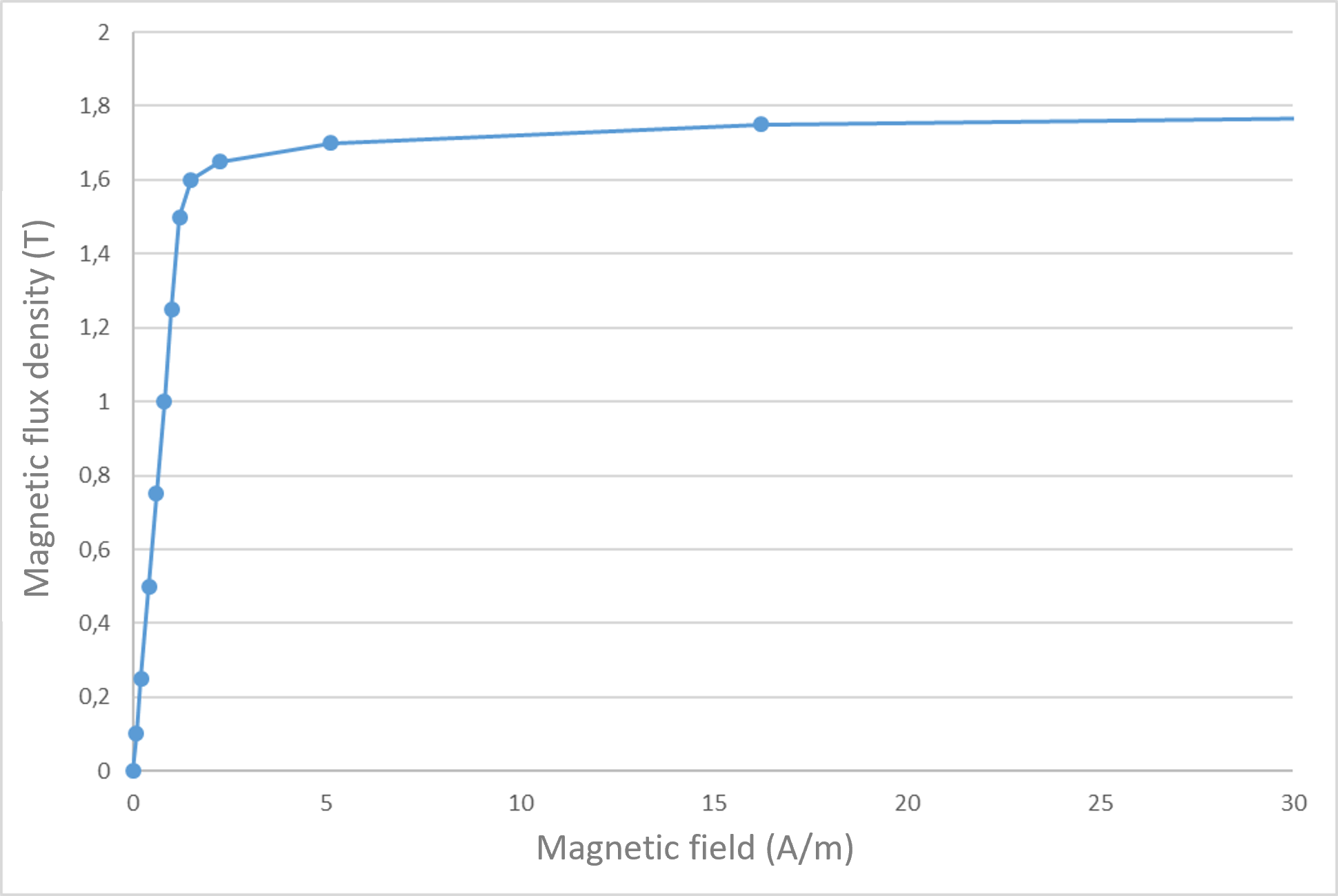}
    \caption{Non-linear behavior law of the transformer core.}
    \label{fig:B_H_transformer}
  \end{figure}

Figure \ref{fig:use-case-transformer} shows the mesh used in the finite elements simulations: grey voxels belong to the core, red voxels to the primary windings, and blue voxels to the secondary windings. The space around the core and windings is also meshed but not represented. In this figure, we can see the three columns of a three-phase transformer, each of them associated with two windings.
A high voltage is imposed in the primary circuit, associated with a high amplitude electric current, which generates a magnetic flux traveling through the core. The core consists of an arrangement of ferromagnetic laminated sheets, associated with a non-linear magnetic anhysteretic permeability as shown in Figure \ref{fig:B_H_transformer}. It will be approximated using the so-called Frohlich model \cite{frohlich1881investigations}: 

\begin{equation}
\mu(H) = \mu_0 + \frac{\alpha}{\beta+ H}
\label{eq:Frohlich}
\end{equation}

In this configuration, we are interested in the magnetic flux density distribution present in the transformer and around it, when the secondary circuit is short-circuited $I_2 = 0$.
The problem is parameterized by the current amplitude $I$ of the primary windings and by the coefficients $\alpha$ and $\beta$ of the Frohlich approximation (equation \ref{eq:Frohlich}).

\textbf{Design of experiments.} Four parameters are considered in this use case: the intensity $I$, the two coefficients $\alpha$ and $\beta$ of the behavior law of the core, and the time $t$. We run 327 simulations, each one corresponding to a parameter set. The values of $I$ are sampled from a beta distribution centered around $1.4 A$, this is done to better learn a known non-linearity physically appearing at this current value. Parameters $\alpha$ and $\beta$ are sampled from a non-informative uniform distribution. The sampling of $t$ is given by the computational code.

\textbf{Why this use case?}
The choice of this use case is motivated by several main arguments. 
First, it is a 3D model initially created for the industrial purpose of design review. Then, applying the proposed method to this use case gives us useful insight, into its capacity to generate accurate results for real-life engineering problems.
Second, the non-linear behavior law of the magnetic core is not often well known by engineers when performing numerical simulations for diagnosis or design review purposes. Then, we aim not merely to evaluate the accuracy of the proposed method on a nonlinear problem, but also to assess the possibility of using this method to solve inverse problems for the deduction of the parameters of the non-linear behavior law. 
Third, in the literature, we can find recent works on surrogate models  \cite{henneron2014model} and deep learning \cite{gong2022further} developments for similar devices, setting a favorable context for future comparison.

\section{Results}
\label{sec:results}

In this section, we jointly discuss the results of the three use cases described in section \ref{sec:use-cases}. We start by presenting, in table \ref{table:POD-energy}, the number of modes necessary to represent the solutions while keeping 95\% , 98\%, and 99\% of the accumulated energy, as specified in equation \ref{eq:sigs}. By comparison with the number of cells given in table \ref{table:case-comparisson}, we observe that the POD (which is the first step of our algorithm) introduces a strong compression on the vector fields associated to the simulation mesh. This can be considered a sort of spatial compression, which is induced by the coding we have chosen for the snapshot matrices.

\begin{table}[ht]
\caption{Comparisson of the two use cases concerning the number of modes necessary to represent the problem, for three levels of cumulated energy (95\%, 98\%, and 99\%).}
\label{table:POD-energy}
\vskip 0.15in
\begin{center}
\begin{small}
\begin{sc}
\begin{tabular}{lcccr}
\toprule
Use Case & \# modes  & \# modes & \# modes\\
 & 95\% & 98\% & 99\% \\
\midrule
Induction   & 2  & 3 & 3  \\
Transformer & 4  & 7 & 10 \\
\bottomrule
\end{tabular}
\end{sc}
\end{small}
\end{center}
\vskip -0.1in
\end{table}

 We calculated, for each of our use cases, the relative RMSE (Root Mean Square Error) and relative AME (Absolute Mean Error). These errors are calculated as follows: 

\begin{equation}
\delta RMSE = \frac{ \frac{1}{N} \sum_{k=1}^{N} \frac{1}{T} \sum_{j=1}^{T} \frac{1}{n} \sum_{i=1}^{n} \| \hat{v} - v \|_{2}^{2}}{ \frac{1}{N} \sum_{k=1}^{N} \frac{1}{T} \sum_{j=1}^{T} \frac{1}{n} \sum_{i=1}^{n}  \| v \|_{2}^{2}}
\label{eq:relRMSE}
\end{equation}

\begin{equation}
\delta AME = \frac{ \frac{1}{N} \sum_{k=1}^{N} \frac{1}{T} \sum_{j=1}^{T} \frac{1}{n} \sum_{i=1}^{n} \mid \hat{v} - v \mid }{ \frac{1}{N} \sum_{k=1}^{N} \frac{1}{T} \sum_{j=1}^{T} \frac{1}{n} \sum_{i=1}^{n} \mid v \mid }
\label{eq:relAME}
\end{equation}

where $\delta$ indicates "relative", $v$ is a reference vector from the test set, $\hat{v}$ is the estimate provided by our method, $N$ is the number of simulations in the design of experiences, $T$ is the number of time steps, and $n$ is the number of mesh cells. We choose to explicitly write three sum signs to remark on the parametric and spatio-temporal nature of the estimated vectorial fields. Table \ref{table:errors} shows these errors, which are multiplied by 100 to express percentages, for the test sets of each use case.

\begin{table}[ht]
\caption{Errors obtained by the surrogate models, on the simulations test sets, for two levels of cumulated energy (95\% and 98\%).}
\label{table:errors}
\vskip 0.15in
\begin{center}
\begin{small}
\begin{sc}
\begin{tabular}{lcccr}
\toprule
Use Case & $\delta$ RMSE & $\delta$ AME & $\delta$ RMSE & $\delta$ AME  \\
 & 95\% &  95\% & 98\% & 98\% \\
\midrule
Induction   & 4.5\% & 4.3\% & 0.98\% & 0.97\% \\
Transformer & 2.9\% & 3.2\% & 2.3\% & 2.5\% \\
\bottomrule
\end{tabular}
\end{sc}
\end{small}
\end{center}
\vskip -0.1in
\end{table}

\begin{figure}[h]
    \centering
    \includegraphics[width=.9\columnwidth]{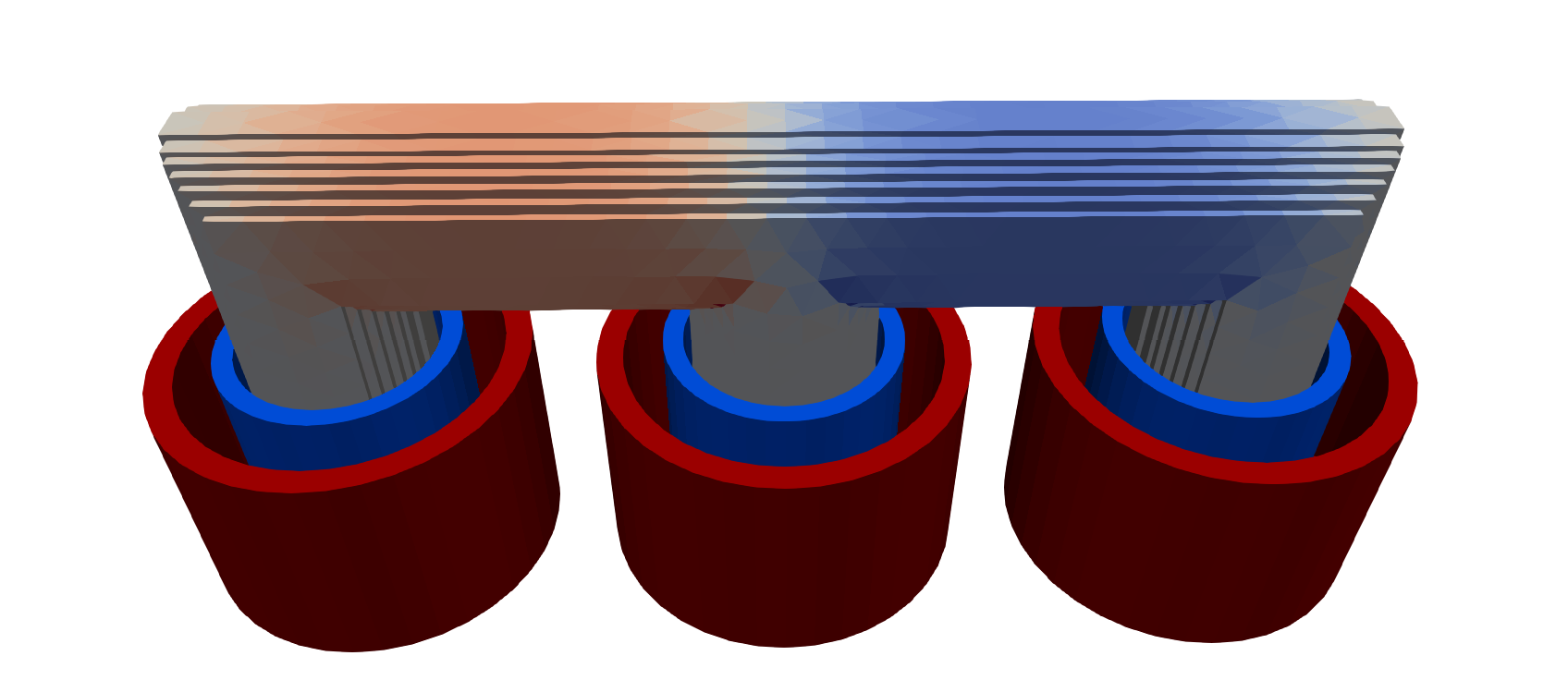}
    \includegraphics[width=.9\columnwidth]{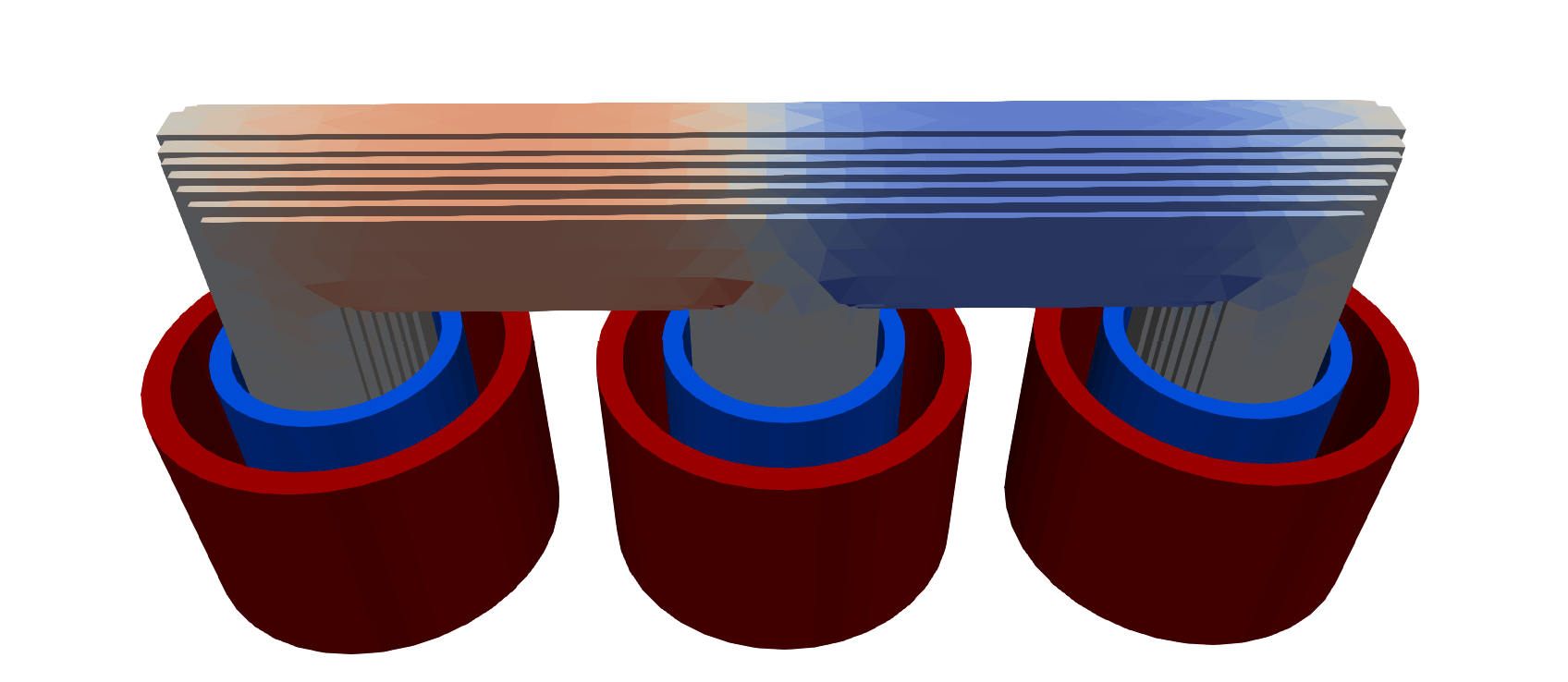}
    \caption{Visual comparison between the modulus of a reconstructed magnetic field (top image) and the reference magnetic field (bottom image), on the test set of the three-phase transformer use case introduced in Section \ref{subsec:use-case-Transfo3D}.}
    \label{fig:view-transformer}
  \end{figure}

Figure \ref{fig:view-transformer} presents a visual comparison between the modulus of a reconstructed magnetic field (top image) and the reference magnetic field (bottom image), on the test set of the three-phase transformer introduced in Section \ref{subsec:use-case-Transfo3D}. For the shown time step and set of parameters, the magnetic field is oscillating on the core. We observe positive values in red, negative in blue, and grey indicates near zero values. We observed that reconstructed and reference images are visually very similar, which was one of the objectives of the proposed surrogate. More importantly, the relative errors shown in Table \ref{table:errors} are very satisfactory.

\section{Real-Time Inference}
\label{sec:realTime}

We recall that our final objective is to develop a surrogate that can be used to power a Digital Twin. This surrogate should be executed on a computer sufficiently fast to allow the interactive exploration of its outputs. 
The introduced method fulfills this objective for several reasons.
First, when coding the snapshot matrix as indicated in equation \ref{eq:snapshot}, the POD acts as a spatial compression system. This means that simulations based on large meshes can potentially be reduced to a vector of relatively few coefficients. Thus the SVRs can perform the estimation of these coefficients very fast. Once the coefficients are obtained, a simple matrix multiplication reconstructs the desired scalar or vectorial field. 
Second, the Direct Time approach avoids iterating overtime at the inference step. 

When designing a visualization system for a Digital Twin,  performing an interactive exploration of the results imposes constraints on the response time of the surrogate model. Response times in human-computer interactions have been studied for years, see for instance \cite{shneiderman1984responseTime}. A classical reference \cite{nielsen1994usability} stated that
0.1 second is about the limit for having the user feel that the system is reacting instantaneously. Thus, we performed an analysis of the response times of the proposed surrogate. For this, we measured the time of inference of a single parameter set, which corresponds to computing a discretized vectorial field for a fixed time step and fixed parameters. We perform this testing in a single node of a cluster, this node contains an Intel-Xeon Platinum 8260 processor operating at 2,40 GHz, and it does not contain GPUs. No special optimization has been performed on the code, composed of Phyton scripts run sequentially. The results are shown in Table \ref{table:times}, times are given in milliseconds. Each shown time is the mean of 100 measurements, the standard deviations are always on the scale $10^5$ thus we consider the measurement reliable. We observe that, for all our use cases the inference time is approximately 2ms, thus around 50 times less (in our slower case) than the time necessary for a fluid real-time user interaction.

\begin{table}[ht]
\caption{Execution time of the surrogate.}
\label{table:times}
\vskip 0.15in
\begin{center}
\begin{small}
\begin{sc}
\begin{tabular}{lcccr}
\toprule
Use Case & \# modes  & \# modes & \\
 & 95\% & 98\%  \\
\midrule
Induction   & 0.43 ms & 0.56 ms \\
Transformer & 1.23 ms & 2.2 ms \\
\bottomrule
\end{tabular}
\end{sc}
\end{small}
\end{center}
\vskip -0.1in
\end{table}

\section{Conclusion}
\label{sec:conclusion}

We have conceived a novel method that combines model reduction via the \textit{Proper Orthogonal Decomposition} (POD) and \textit{Support Vector Regression} (SVR). The aim is the construction of a learning-based surrogate model for parameterized  PDEs, which are also temporal and spatial processes. We found an error upper bound of the proposed method and a mathematical proof of this bound is included. 
%This method is non-invasive and uses a direct-time estimation strategy. 
We have performed tests on two use cases
concerning electrical machines.
These use cases are not toy examples. They are produced by an industrial computational code, they use 3D meshes representing non-trivial geometries and contain non-linearities. The obtained results show a good reconstruction accuracy. Furthermore, we have studied the response time of the proposed surrogate. The tests indicate that it is adapted to real-time tasks.

The current results indicate that the method can be applied to the interactive exploration of use cases based on much larger meshes or for higher intrinsic complexity. Indeed, the slower case we treated was 50 times faster than needed for interactive exploration. Moreover, no code optimization has been performed yet. Thus, the proposed method presents great potential for building parameterized surrogates of industrial-level numerical simulations.

\bibliography{bibliography}
\bibliographystyle{icml2023}

%%%%%%%%%%%%%%%%%%%%%%%%%%%%%%%%%%%%%%%%%%%%%%%%%%%%%%%%%%%%%%%%%%%%%%%%%%%%%%%
%%%%%%%%%%%%%%%%%%%%%%%%%%%%%%%%%%%%%%%%%%%%%%%%%%%%%%%%%%%%%%%%%%%%%%%%%%%%%%%
% APPENDIX
%%%%%%%%%%%%%%%%%%%%%%%%%%%%%%%%%%%%%%%%%%%%%%%%%%%%%%%%%%%%%%%%%%%%%%%%%%%%%%%
%%%%%%%%%%%%%%%%%%%%%%%%%%%%%%%%%%%%%%%%%%%%%%%%%%%%%%%%%%%%%%%%%%%%%%%%%%%%%%%
\newpage
\appendix
\onecolumn

\newcommand*\circled[1]{\tikz[baseline=(char.base)]{
            \node[shape=circle,draw,inner sep=1.5pt] (char) {#1};}}
            
\section{Appendix: inference error upper bound}
\label{sec:appendix1}

\subsection{Notation}

\begin{itemize}
    \item Let $X \in \mathbb{R}^{n\times m}$ be a snapshot matrix and  $\hat{X}$ its inferred equivalent.
    \item  Let  $C, \hat{C} \in \mathbb{R}^{r\times m}$ be their coefficient matrices which are the projection of $X, \hat{X}$ by $U^r_{POD} \in \mathbb{R}^{r\times n} $. They verify :
    \begin{equation}
    \begin{aligned}
    & X=U_{POD}^r.C \\
    & \hat{X}=U_{POD}^r.\hat{C}
    \end{aligned}
    \end{equation}

    \item We denote $\mathbf{X}_p=(X_{p1},...,X_{pm})^{\top}$ ,  $\hat{\mathbf{X}}_p=(\hat{X}_{p1},...,\hat{X}_{pm})^{\top}$ the snapshot values of $X$ and $\hat{X}$ at the cell $p\leq n$.
    \item Similarly, we define  $\mathbf{C}_k=(C_{k1},...,C_{km})^{\top}$ and $\hat{\mathbf{C}}_k=(\hat{C}_{k1},...,\hat{C}_{km})^{\top}$ the k-th component of the original and the inferred matrices for any $k\leq r$.
    \item  By centering and reducing $\mathbf{C}_k=(C_{k1},...,C_{km})^{\top}$, we derive $\mathbf{c}_k$ and $\hat{\mathbf{c}}_k$ which correspond to the training vector and the output of the k-th SVR respectively. They verify :
    \begin{equation}
    \begin{aligned}
    & \mathbf{C}_k=S_k.\mathbf{c}_k+E_k \\
    & \hat{\mathbf{C}}_k=S_i.\hat{\mathbf{c}}_k+E_k
    \end{aligned}
    \end{equation}
    where $E_k$ is the mean of $\mathbf{C}_k$ and $S_k$ its standard deviation.
    
\end{itemize}

\subsection{Introduction}
\newtheorem{inequality}{Inequality}[theorem]

SVR machines are trained with respect to the \textit{Root Mean Square Error} of inference $\mathbf{e}_k$ evaluated between the output $\hat{\mathbf{c}}_k$ and the standardized k-th component of the original coefficient matrix $\mathbf{c}_k$. Thus:

\begin{equation*}
\mathbf{e}_k= \| \mathbf{c}_k-\hat{\mathbf{c}}_k \|_{2}    
\end{equation*}\\

To optimize the learning hyper-parameters $\lambda_1,\lambda_2,...$ of the metamodel, we aim to minimize an objective function $\mathbf{e}$, equal to the highest of validation errors $(\mathbf{e}_1,\mathbf{e}_2,...,\mathbf{e}_r)$ recorded among all $\mathbf{r}$ SVR machines such as :\\

\begin{equation}
\mathbf{e}= \max_{1\leq k \leq r}{\mathbf{e}_k}    
\end{equation}\\

In fact, we can prove that the root mean square error of inference of $X$ computed at any cell $p \leq n$ has an upper bound that is linearly proportional to $\mathbf{e}$ (\textit{Theorem A.1}).

\begin{theorem}
    
\(\forall p \leq n \)      \(\exists K_p >0 \)  that verifies 
\[ \| \mathbf{X}_p-\hat{\mathbf{X}}_p \|_2 \leq  K_p.\mathbf{e} \]
\end{theorem}

Thus, minimizing the objective function contributes to shrinking the validation error evaluated at any component $p\leq n$ of the snapshot matrix.\\

\subsection{Proof}

At first, we remind the analytical formulation of $\| X_p-\hat{X}_p \|_2$ : \\

$$\| X_p-\hat{X}_p \|_2^2=\frac{1}{m}\sum_{j=1}^{j=m}(X_{pj}-\hat{X}_{pj})^2$$\\

To find an upper bound, we start by expressing the relative error of inference $\Delta X_{pj}=X_{pj}-\hat{X}_{pj}$ of the $j-th$ snapshot at the cell $p$ in terms of inference errors of all SVR machines.\\

$ \forall p \in [1,N] \ \   \forall j\in[1,m] \ :$ 

\begin{equation*}\label{eq1}
\begin{aligned}
\Delta X_{pj}&=X_{pj}-\hat{X}_{pj}\\
(1) \implies &=\sum_{k=1}^{k=r}U_{POD_{pk}}^r(C_{kj}-\hat{C}_{kj})\\
(2) \implies&=\sum_{k=1}^{k=r}U_{POD_{pk}}^r(S_k c_{kj}+E_k-S_k \hat{c}_{kj}-E_k)\\
&=\sum_{k=1}^{k=r}S_k.U_{POD_{pk}}^r\underbrace{(c_{kj}-\hat{c}_{kj})}_{\Delta c_{kj}}\\
\end{aligned}
\end{equation*}

\begin{equation}
    \begin{aligned}
        \implies \Delta X_{pj}&=\sum_{k=1}^{k=r}S_k.U_{POD_{pk}}^r\Delta c_{kj}\\
    \end{aligned}
\end{equation}\\

By squaring $\Delta X_{pj}$, we obtain : \\

\begin{equation*}
\begin{aligned}
\Delta X_{pj}^2&=(\sum_{k=1}^{k=r}S_k. U_{POD_{pk}}^r\Delta c_{kj})^2\\
&=\sum_{k=1}^{k=r}(S_k. U_{POD_{pk}}^r)^2 \Delta c_{kj}^2+2\sum_{1\leq h<l \leq r}(S_h. U_{POD_{ph}}^r).(S_l. U_{POD_{pl}}^r).\Delta c_{hj}\Delta c_{lj}\\
\Delta X_{pj}^2&\leq\sum_{k=1}^{k=r}(S_k. U_{POD_{pk}}^r)^2 \Delta c_{kj}^2+2\sum_{1\leq h<l \leq r}\lvert S_h. U_{POD_{ph}}^r\rvert . \lvert S_l. U_{POD_{pl}}^r\rvert . \lvert \Delta c_{hj}\Delta c_{lj}\rvert\\
\end{aligned}
\end{equation*}\\

Then, we sum over $j$:\\

\begin{equation*}
\begin{aligned}
\sum_{j=1}^{j=m}\Delta X_{pj}^2&\leq \sum_{j=1}^{j=m}\sum_{k=1}^{k=r}(S_k. U_{POD_{pk}}^r)^2 \Delta c_{kj}^2+2\sum_{j=1}^{j=m}\sum_{1\leq h<l \leq r}\lvert S_h. U_{POD_{ph}}^r\rvert . \lvert S_l. U_{POD_{pl}}^r\rvert . \lvert \Delta c_{hj}\Delta c_{lj}\rvert\\
&\leq \sum_{k=1}^{k=r}(S_k. U_{POD_{pk}}^r)^2 \underbrace{\sum_{j=1}^{j=m} \Delta c_{kj}^2}_{\circled{1}}+2\sum_{1\leq h<l \leq r}\lvert S_h. U_{POD_{ph}}^r\rvert . \lvert S_l. U_{POD_{pl}}^r\rvert  \underbrace{\sum_{j=1}^{j=m} \lvert \Delta c_{hj}\Delta c_{lj}\rvert}_{\circled{2}}\\
\end{aligned}
\end{equation*}\\

We recall that $\forall k \in [1,r]$ : 
\begin{equation}
    \begin{aligned}
  \mathbf{e}_k^2= \|\mathbf{c}_k-\hat{\mathbf{c}}_k\|_2^2=\dfrac{1}{m}\sum_{j=1}^{j=m}\Delta c_{kj}^2      
    \end{aligned}
\end{equation}\\

Thus, term $\circled{1}$ immediately becomes $\sum_{j=1}^{j=m}\Delta c_{kj}^2=m\mathbf{e}_k^2 \leq m\mathbf{e}^2$.

By virtue of Hölder's inequality and $(5)$, we establish the following about term $\circled{2}$: \\
\begin{equation*}
\begin{aligned}
  \sum_{j=1}^{j=m}\lvert\Delta c_{hj}\Delta c_{lj}\rvert &\leq \left( \sum_{j=1}^{j=m}\Delta c_{hj}^2\right)^{1/2}\cdot \left(\sum_{j=1}^{j=m}\Delta c_{lj}^2\right)^{1/2}\\
   &\leq (m\mathbf{e}_h^2)^{1/2}\cdot(m\mathbf{e}_l^2)^{1/2}\\
   &\leq m\mathbf{e}_h\cdot \mathbf{e}_l\leq m\mathbf{e}^2
\end{aligned}    
\end{equation*}

By introducing the new bounds on term $\circled{1}$ and $\circled{2}$, the latter inequality becomes : \\

\begin{equation*}
\begin{aligned}
\sum_{j=1}^{j=m}\Delta X_{pj}^2&\leq \sum_{k=1}^{k=r}(S_k. U_{POD_{pk}}^r)^2 \cdot (m\mathbf{e}^2)+2\sum_{1\leq h<l \leq r}\lvert S_h. U_{POD_{ph}}^r\rvert . \lvert S_l. U_{POD_{pl}}^r\rvert  \cdot (m\mathbf{e}^2)\\
&\leq \left( \sum_{k=1}^{k=r}(S_k. U_{POD_{pk}}^r)^2 +2\sum_{1\leq h<l \leq r}\lvert S_h. U_{POD_{ph}}^r\rvert . \lvert S_l. U_{POD_{pl}}^r\rvert \right) \cdot m\mathbf{e}^2   \\
\end{aligned}
\end{equation*}

\begin{equation*}
\begin{aligned}
\frac{1}{m}\sum_{j=1}^{j=m}\Delta X_{pj}^2&\leq \left( \sum_{k=1}^{k=r}(S_k. U_{POD_{pk}}^r)^2 +2\sum_{1\leq h<l \leq r}\lvert S_h. U_{POD_{ph}}^r\rvert . \lvert S_l. U_{POD_{pl}}^r\rvert \right) \cdot \mathbf{e}^2   \\
\end{aligned}
\end{equation*}

\begin{equation*}
\begin{aligned}
\|\mathbf{X}_p-\mathbf{X}_p\|_2&\leq \left( \sum_{k=1}^{k=r}(S_k. U_{POD_{pk}}^r)^2 +2\sum_{1\leq h<l \leq r}\lvert S_h. U_{POD_{ph}}^r\rvert . \lvert S_l. U_{POD_{pl}}^r\rvert \right)^{1/2} \cdot \mathbf{e}   \\
\end{aligned}
\end{equation*}\\

Hence :

    \[ \|\mathbf{X}_p-\mathbf{X}_p\|_2\leq K_p \cdot \mathbf{e} \] where \(K_p =  \left( \sum_{k=1}^{k=r}(S_k. U_{POD_{pk}}^r)^2 +2\sum_{1\leq h<l \leq r}\lvert S_h. U_{POD_{ph}}^r\rvert . \lvert S_l. U_{POD_{pl}}^r\rvert \right)^{1/2} \)

% \input{appendix2}

%%%%%%%%%%%%%%%%%%%%%%%%%%%%%%%%%%%%%%%%%%%%%%%%%%%%%%%%%%%%%%%%%%%%%%%%%%%%%%%
%%%%%%%%%%%%%%%%%%%%%%%%%%%%%%%%%%%%%%%%%%%%%%%%%%%%%%%%%%%%%%%%%%%%%%%%%%%%%%%

\end{document}